\documentclass[runningheads]{llncs}
\usepackage{graphicx}
\usepackage{amssymb}
\usepackage{adjustbox}
\usepackage{hyperref}
\hypersetup{colorlinks=true,
urlcolor=blue}
\begin{document}
%
%\title{Speech-Driven 3D Talking Heads using Facial Landmarks}
%\title{3D Talking Heads Animation Using Facial Landmarks}
%\title{Landmarks-based Animation of Identity-Agnostic 3D Talking Heads}
%\title{Landmarks-Based Identity-Agnostic 3D Talking Heads Generation}
\title{Learning Landmarks Motion from Speech for Speaker-Agnostic 3D Talking Heads Generation}
\titlerunning{Speech-Driven 3D Talking Heads Generation}
\author{Federico Nocentini\inst{1}\orcidID{0009-0000-6585-6924} \and
Claudio Ferrari\inst{2}\orcidID{0000-0001-9465-6753} \and
Stefano Berretti\inst{1}\orcidID{0000-0003-1219-4386}}
\authorrunning{F. Nocentini et al.}
%\authorrunning{Author et al.}
\institute{Media Integration and Communication Center (MICC), Università degli Studi di Firenze \\
\email{\{federico.nocentini,stefano.berretti\}@unifi.it}\\\and
Department of Engineering and Architecture, Università degli studi di Parma \\
\email{claudio.ferrari2@unipr.it}\\}
\maketitle              % typeset the header of the contribution
\begin{abstract}
This paper presents a novel approach for generating 3D talking heads from raw audio inputs. Our method grounds on the idea that speech related movements can be comprehensively and efficiently described by the motion of a few control points located on the movable parts of the face, \textit{i.e.}, landmarks. The underlying musculoskeletal structure then allows us to learn how their motion influences the geometrical deformations of the whole face. The proposed method employs two distinct models to this aim: the first one learns to generate the motion of a sparse set of landmarks from the given audio. The second model expands such landmarks motion to a dense motion field, which is utilized to animate a given 3D mesh in neutral state. Additionally, we introduce a novel loss function, named Cosine Loss, which minimizes the angle between the generated motion vectors and the ground truth ones. Using landmarks in 3D talking head generation offers various advantages such as consistency, reliability, and obviating the need for manual-annotation. Our approach is designed to be identity-agnostic, enabling high-quality facial animations for any users without additional data or training. Code and models are available at: \href{https://github.com/FedeNoce/s2l-s2d}{S2L+S2D}
\end{abstract}
\section{Introduction}
Speech-driven 3D talking heads generation is a rapidly growing field of research and development that has garnered significant interest in recent years. This technology involves generating realistic 3D digital avatars that can accurately replicate human speech and facial expressions. This innovation has far-reaching implications for a wide range of applications, including virtual assistants, video games, education, and entertainment.
One of the most significant advantages of speech-driven 3D talking heads is the ability to create immersive and engaging user experiences.
This technology can be used to enhance communication in many different domains, from customer service to online education, and can provide a more human-like interaction than traditional text or voice-only interfaces.
Furthermore, speech-driven 3D talking heads can have a significant impact on accessibility and inclusivity. By providing a visual representation of speech and language, this technology can help individuals with hearing or speech impairments to communicate more effectively. 
Recent advancements in speech-driven 3D facial animation have focused on two primary approaches: vertex-based animation and parameter-based animation.
Vertex-based approaches utilize mappings from audio to sequences of 3D face models, with mesh vertex positions predicted to animate the model. However, the main challenge of this approach lies in the complexity of the resulting models, as they must learn to generate vertex mesh sequences containing a large number of 3D points.
Parameter-based approaches generate animation curves from audio, resulting in sequences of animation parameters. However, a significant challenge of this approach is converting a sequence of 3D meshes into a sequence of parameters, typically requiring hand-annotated viseme or blendshapes as a starting point.

\begin{figure}[h!]
\includegraphics[width=\textwidth]{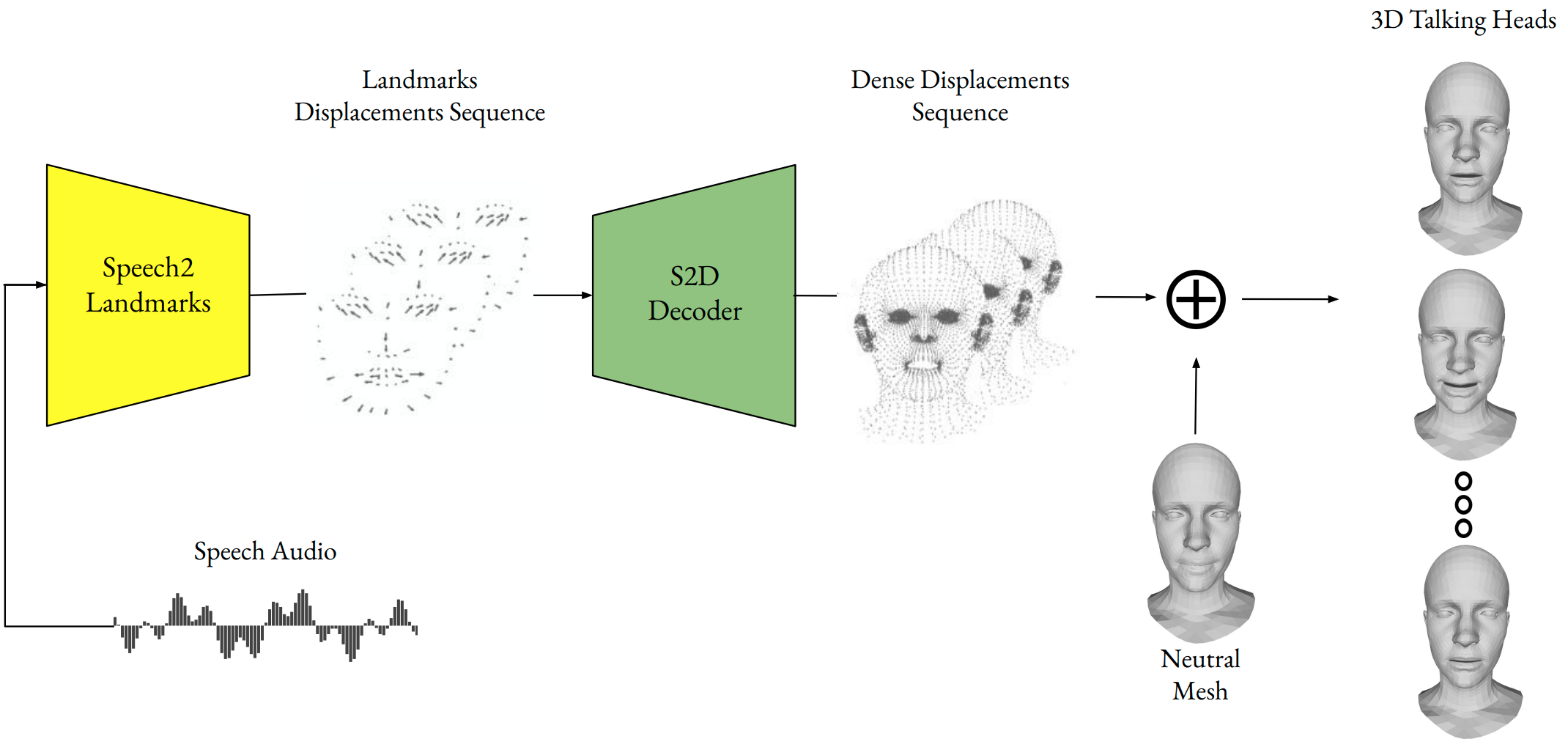}
\caption{Overview of our architecture: given a raw audio input and a neutral 3D face mesh as input, our proposed approach can synthesize a sequence of realistic 3D facial motions with precise lip movements.} \label{fig:overview}
\end{figure}

In this paper, we introduce a novel approach for generating 3D talking heads that decomposes the problem into two distinct sub-problems, each tackled by a separate model, as described in Figure~\ref{fig:overview}. 
The first model tracks the movements of scattered landmarks in response to the speech. Specifically, it takes an audio signal as input, from which it generates a frame-by-frame motion of a set of landmarks. The motion is modeled as displacement relative to a neutral configuration of 3D landmarks. 
The second model takes the resulting displacement of scattered landmarks and densifies them to create a dense motion field. Using the latter, the model then animates a 3D face mesh by adding the motion field to the 3D face vertices. 
%Our approach can be viewed as a hybrid method that combines parameter-based and vertex-based techniques, as the displacements of the landmarks with respect to the neutral configuration can be considered as a parameterization of the mesh.
By addressing each sub-problem independently, we aim to improve the overall performance and efficiency of our approach for generating high-quality 3D talking heads.
The use of landmark displacements to model speech movements in 3D talking head generation offers several key advantages:
firstly, the use of landmarks provides a consistent and reliable way to define the structure of the face, which makes it easier to generate realistic facial expressions. 
Secondly, landmarks displacements can be interpreted as parameters, eliminating the need for hand-annotation.
Thirdly, training the model to predict landmarks displacements from audio allows for complete independence from the identity of the speaker. This enables the predicted displacements from a given audio to be used for animating multiple identities without requiring the model to be retrained.
Finally, landmarks are particularly effective for representing the movement of the mouth during speech, making them an ideal choice for speech-driven 3D talking heads.

\section{Related Works}
In the following, we summarize the work in the literature that are closer to our proposed solution distinguishing between 2D and 3D methods.
Several previous studies have focused on the generation of 2D talking head videos driven by speech. Suwajanakorn et al.~\cite{10.1145/3072959.3073640} utilized an LSTM network trained on 19 hours of video footage of former President Obama to predict his specific 2D lip landmarks from speech inputs, which was then used for image generation. 
Vougioukas et al.~\cite{DBLP:journals/corr/abs-1906-06337} proposed a method for generating facial animation from a single RGB image using a temporal generative adversarial network. 
Chung et al.~\cite{DBLP:journals/corr/ChungJZ17} introduced a real-time approach for generating an RGB video of a talking face by directly mapping audio input to the video output space, which can be used to redub a new target identity not seen during training. 
Landmarks have also emerged as a powerful tool for generating 2D talking heads from speech inputs~\cite{DBLP:journals/corr/abs-2004-12992}. By furnishing a concise encoding of facial motion, these landmarks can be reliably estimated through computer vision methodologies. Nonetheless, their usefulness is confined to 2D rendering and fail to account for the comprehensive 3D structure of the face. In this study, we address this inadequacy of 2D landmark extraction by employing 3D landmarks obtained from meshes.
Methods for 2D talking head generation focus on generating realistic lip motion, posing less attention to the geometrical face deformations which are mainly induced by texture changes. In the 3D domain instead, facial deformations are to be accounted from the geometric perspective. 
In earlier attempts, researchers concentrated on animating a pre-designed facial rig with the aid of procedural rules. For instance, HMM-based models generated visemes from input audio or text, and the ensuing facial animations were generated with viseme-dependent co-articulation models or through blending facial templates~\cite{DEMARTINO2006971,10.1145/2897824.2925984,982373}. In particular, these methods are based on pre-trained speech models to create an abstract and generalized representation of the audio input. A CNN or autoregressive model then interprets this representation to map it either to a 3DMM space or directly to 3D meshes. For example, Karras et al.~\cite{10.1145/3072959.3073658} learned a 3D facial animation model from 3-5 minutes of high-quality actor-specific 3D data. Similarly, VOCA~\cite{Cudeiro_2019_CVPR} is trained on 3D data of multiple subjects and can animate the corresponding set of identities from input audio. Meanwhile, MeshTalk~\cite{DBLP:journals/corr/abs-2104-08223} learns a categorical representation for facial expressions and auto-regressively samples from this categorical space to animate a given 3D facial template mesh of a subject from audio inputs. FaceFormer~\cite{DBLP:journals/corr/abs-2112-05329}, on the other hand, uses a transformer-based decoder to regress displacements on top of a template mesh.
While both VOCA and FaceFormer require a speaker identification code for the model to choose from the training set's talking styles, our approach differs in that it is completely identity independent.
%Moreover, the computational complexity of the transformer-based decoder in Faceformer prohibits real-time sequence prediction, necessitating a reduction in the generated sequence's frame rate.
%In order to mitigate these challenges, we propose the utilization of a computationally less burdensome model, thereby enabling us to generate sequences with higher framerates and at a much faster pace.
%In order to speed-up the animation process, we propose the utilization of a computationally less burdensome model, thereby enabling us to generate sequences with higher framerates and at a faster pace.
In contrast to existing methods, our work aims to predict 3D facial animations from speech that can be used to animate 3D digital avatars independently of the speaker's identity.

\section{Proposed Approach}
We propose a novel method to generate 3D facial animations using only an audio input. Unlike existing methods, our framework is designed to be completely agnostic to the identity of the subject being animated. Our approach involves two separately trained models that work in tandem to generate realistic and expressive facial animations from audio inputs. Through the decoupling of landmarks displacement generation and densification, our method enables the generation of high-quality facial animations for any user without the need for additional data or training.
Our proposed methodology deviates from the existing literature by not directly synthesizing meshes from audio input. This novel approach offers the advantage of reducing the computational complexity of speech motion generation as the number of landmarks is significantly fewer than that of mesh vertices. Additionally, the independence of the two models allows for the generation of landmarks displacements that can animate a variety of meshes.

\subsection{Speech2Landmarks (S2L)}
Let $L = \left \{(L_i^{gt}, L_i^n, A_i)\right \}_{i=0}^N$ denote the training set comprising $N$ samples, where $A_i$ is an audio containing a spoken sentence, $L_i^{gt} = (l_i^0, \dots l_i^{K_i}) \in \mathbb{R}^{K_i \times 68 \times 3}$ represents the facial landmark sequence of length $K_i$ that corresponds to the spoken sentence in audio $A_i$ and $L_i^n\in \mathbb{R}^{68 \times 3}$ are the landmarks of the neutral face.
To derive the landmarks displacement dataset with respect to the neutral configuration, we apply a transformation that results in $L_d = \left \{(S_i^{gt}, L_i^n, A_i)\right \}_{i=0}^N$, where $S_i^{gt} = (s_i^0, \dots s_i^{K_i}) \in \mathbb{R}^{K_i \times 68 \times 3}$ and each $s_i^k = l_i^k - L_i^n \in \mathbb{R}^{ 68 \times 3}$ is the landmarks displacements.
Our aim is to learn a mapping function (S2L) that establishes a correspondence between the audio input $A_i$ and the ground-truth landmark displacements $S_i^{gt}$, which is realized by assembling a three-part composite model comprising a Wav2Vec Encoder, a multilayer bidirectional LSTM, and a fully connected layer.
The utilization of a pre-trained audio processing model enhances the generalization capabilities of our framework, thereby enabling us to animate sentences in languages other than the one on which the model was trained. This approach expands the versatility and applicability of our framework.

\subsubsection{Wav2Vec} Our method employs a generalized speech model to encode audio inputs $A$. Specifically, we use the Wav2Vec 2.0 model~\cite{DBLP:journals/corr/abs-2006-11477}, which is based on a CNN architecture trained in a self-supervised and semi-supervised manner to produce a meaningful latent representation of human speech. To enable learning from a large amount of unlabeled data, the model is trained using a contrastive loss. Wav2Vec 2.0 extends this architecture by incorporating a Transformer-based architecture~\cite{NIPS2017_3f5ee243} and quantizing the latent representation. To match the sampling frequency of the motion (60fps for the VOCAset with 16kHz audio), we resample the Wav2Vec 2.0 output using a linear interpolation layer, resulting in a contextual representation:
\begin{equation}
A \rightarrow 	\left \{ a_i \right \}_{i=0}^T .
\end{equation}

\noindent
where $T$ is the number of frames extracted from the audio. In this study, we utilize a pre-trained version of the Wav2Vec 2.0 encoder.

%\subsubsection{Multilayer Bi-LSTM + FC}
%Multilayer Bidirectional Long Short-Term Memory (Bi-LSTM) is a deep learning architecture that has gained significant attention in recent years for its ability to model sequential data. Bi-LSTM is a variant of the recurrent neural network (RNN) that utilizes both forward and backward hidden states to capture the dependencies in the input sequence. The multilayer extension of Bi-LSTM involves stacking multiple Bi-LSTM layers to increase the model's capacity and improve its ability to learn complex representations of the input.
%Our proposed approach leverages the contextualized representation acquired via the Wav2Vec model, and employs a concatenation scheme involving a multilayer bidirectional LSTM and a fully-connected layer to produce the landmarks shift sequence that corresponds to the audio input.

%\begin{equation}
%D = S2S\left ( A\right )
%\end{equation}

\subsection{Sparse2Dense (S2D)}
Otbertout et al.~\cite{otberdout2022sparse,otberdout2023generating} presented the S2D Decoder, which is based on the spiral operator proposed in~\cite{DBLP:journals/corr/abs-1905-02876}. In the following, all meshes employed possess a uniform topology and are in complete point-to-point correspondence. The training set $L = \left \{(M_i^{n}, M_i^{gt}, L_i^n, L_i^{gt})\right \}_{i=0}^N$ consists of $N$ samples, where $M_i^{n} \in \mathbb{R}^{M \times 3}$ represents a neutral 3D face, $M_i^{gt} \in \mathbb{R}^{M \times 3}$ represents a 3D talking head, and $L_i^n \in \mathbb{R}^{68 \times 3}$ and $L_i^{gt} \in \mathbb{R}^{68 \times 3}$ denote the 3D landmarks that correspond to $M_i^{n}$ and $M_i^{gt}$, respectively.
To generate the sparse-to-dense displacement dataset, we employ a transformation that yields $L_d = \left \{(D_i, s_i)\right \}_{i=0}^N$, where $D_i = M_i^{gt} - M_i^{n}$ and $s_i = L_i^{gt} - L_i^{n}$. 
The S2D Decoder takes the landmarks' displacements as input and produces the corresponding 3D mesh vertex displacements. This model transforms a sparse set of scattered displacements into a dense set of displacements by utilizing five spiral convolution layers, each of which is followed by an up-sampling layer. In order to obtain the reconstructed mesh utilizing the model prediction, we employ the following equation: $\hat{M_i} = \hat{D_i} + M_i^n$, where $\hat{D_i}$ represents the model prediction and $M_i^n$ represents the mesh in its neutral expression.
\section{Training}
In order to accelerate the training process, we opted to train the two models independently. This approach resulted in improved convergence for both models. 
Both models were trained on the same training set of VOCAset~\cite{Cudeiro_2019_CVPR}, which comprises paired audio phrases and 3D talking head animations.

\subsection{S2L Losses}
For the training of our S2L model, we formulated a loss function comprising of four terms, which can be expressed as:
\begin{equation}
L_{S2L} = \lambda_1 L_{rec} + \lambda_2 L_{mouth} + \lambda_3 L_{cos} + \lambda_4 L_{vel}.
\end{equation}

\noindent
Here, $L_{rec}$ represents the loss incurred in reconstructing the facial landmark displacements, while $L_{mouth}$ corresponds to the loss incurred in reconstructing the mouth landmark displacements. Additionally, $L_{vel}$ denotes the velocity loss, and $L_{cos}$ signifies the cosine loss. The hyperparameters $\lambda_1$, $\lambda_2$, $\lambda_3$, and $\lambda_4$ control the contribution of each loss term in the overall loss function.

\subsubsection{Reconstruction Losses:}
The reconstruction loss $L_{rec}$ is defined as the $L_2$ norm computed between all generated landmarks displacements and their respective ground truth counterparts. Specifically, this loss function is applied uniformly across all generated landmarks displacements. In a similar vein, the reconstruction loss for the mouth region $L_{mouth}$ is formulated as an $L_2$ norm, with the exception that it is only calculated for the landmarks displacements that are more important during speech, namely those of the mouth and jaw:
\begin{equation}
L_{rec} = \frac{1}{N}\sum\limits_{n=1}^N \frac{1}{T_n}\sum\limits_{t=1}^{T_n} \left \|d_{n,t} - \hat{d}_{n,t} \right \|_2 ,
\end{equation}

\begin{equation}
L_{mouth} = \frac{1}{N}\sum\limits_{n=1}^N \frac{1}{T_n}\sum\limits_{t=1}^{T_n} \Bigl \|m_{n,t} - \hat{m}_{n,t} \Bigr \|_2 .
\end{equation}

\noindent
where $N$ refers to the number of sequences, $T_n$ corresponds to the length of the $n^{th}$ sequence, while $d_{n,t}$ and $m_{n,t}$ represent the respective ground truth values for all displacements and mouth/jaw displacements. Conversely, $\hat{d}_{n,t}$ and $\hat{m}_{n,t}$ denote the model predictions for all displacements and mouth/jaw displacements, respectively.
In order to enhance the convergence of our displacement prediction model, we propose the use of a cosine loss. By incorporating this loss, we aim to minimize the angle between predicted and ground truth displacements, thereby improving the overall performance of the model:
\begin{equation}
L_{cos} = \frac{1}{N}\sum\limits_{n=1}^N \frac{1}{T_n}\sum\limits_{t=1}^{T_n}1 - \frac {d_{n,t} \cdot \hat{d}_{n,t}} { \Bigl \|d_{n,t} \Bigr \|_{2} \left \| \hat{d}_{n,t}\right \|_{2}} .
\end{equation}

\subsubsection{Temporal consistency loss:}
In our efforts to augment the temporal consistency of our model, we use a loss, denoted as velocity loss and inspired by~\cite{Cudeiro_2019_CVPR}, that aims to minimize the $L_2$ norm of the pairwise differences. 
\begin{equation}
L_{vel} = \frac{1}{N}\sum\limits_{n=1}^N \frac{1}{T_n}\sum\limits_{t=2}^{T_n}\left \|(d_{n,t} - d_{n,t-1}) - (\hat{d}_{n,t} - \hat{d}_{n,t-1}) \right \|_2 .
\end{equation}

\subsection{S2D Losses}
In order to enhance the efficacy of S2D Decoder training, we propose a three-loss framework. This framework comprises of three unique losses. The initial two losses are similar to those expounded earlier for S2L and function directly on the displacements. Meanwhile, the third loss governs the precision of the generated mesh. Again, the hyperparameters $\lambda_5$, $\lambda_6$ and $\lambda_7$ control the contribution of each loss term in the overall loss function. We define the loss as follows:
\begin{equation}
L_{S2D} = \lambda_5 L_{rec} + \lambda_6 L_{cos} + \lambda_7 L_{weighted} .
\end{equation}

\subsubsection{Reconstruction Loss:} 
The dense displacement reconstruction is subject to a reconstruction loss, which is defined as follows:
\begin{equation}
L_{rec} = \frac{1}{N}\sum\limits_{i=1}^N \left \|D_{n} - \hat{D}_{n} \right \|_2 .
\end{equation}

\noindent
where $\hat{D}_{n}$ denotes the predicted value by the model, and ${D}_{n}$ represents the corresponding ground truth.

\subsubsection{Weighted Loss:}
To enhance reconstruction accuracy, we introduce an additional loss term that minimizes the discrepancy between the estimated shape $\hat{M_i}$ and the actual expressive mesh $M_i$. Notably, the vertices in close proximity to the landmarks are susceptible to more significant deformations, while other regions, such as the forehead, remain relatively stable. Thus, similar to~\cite{otberdout2022sparse}, we propose a weighted $L_2$ loss, where certain regions of the mesh are assigned more weight to account for their greater importance in the reconstruction process:
\begin{equation}
L_{weighted} = \frac{1}{N}\sum\limits_{i=1}^N w_i \Bigl \|p_{i} - \hat{p}_{i} \Bigr \|_2 .
\end{equation}

\noindent
Following~\cite{otberdout2022sparse}, we use a specific method for defining the weights, denoted as $w_i$, on a mesh represented by vertices $p_i$ and landmarks $l_j$. Specifically, the weight of each vertex is defined as the inverse of the Euclidean distance between the vertex and its closest landmark, i.e., $w_i = \frac{1}{min\ d(p_i, l_j)}, \forall j$. This weighting scheme provides a coarse estimation of the contribution of each vertex $p_i$ to the generation of lip movements. Given that the mesh topology is constant, we precompute the weights $w_i$ and leverage them across all samples.

\subsection{Training Details}

The S2L model was trained using the Adam optimizer for 300 epochs with a learning rate of $10^{-4}$. The model's bi-directional Long Short-Term Memory (Bi-LSTM) architecture comprises three layers, each with a hidden size of 64. The loss function, $L_{S2L}$, includes four regularization terms, namely, $\lambda_1=10^{-1}$, $\lambda_2=1$, $\lambda_3=10^{-4}$, and $\lambda_4=10$.
On the other hand, the S2D model was trained using the Adam optimizer for 300 epochs with a learning rate of $10^{-3}$. The model is built by concatenating five spiral convolution layers with an upsampling layer. The loss function, $L_{S2D}$, includes three regularization terms, namely, $\lambda_5=10^{-1}$, $\lambda_6=10^{-4}$, and $\lambda_7=1$.

\subsection{Inference Time}
To obtain the talking heads after the training of the models, the following steps are followed:
\begin{enumerate}
  \item \textbf{S2L} takes an audio file $A_i$ as input and produces a sequence of landmark displacements denoted as $\hat{S_i} = S2L(A_i) = \left(\hat{s_i}^0, \dots, \hat{s_i}^{K_i}\right) \in \mathbb{R}^{K_i \times 68 \times 3}$.
  \item \textbf{S2D} takes the landmarks displacement generated by \textbf{S2L}, denoted as $\hat{S_i}$, and produces a sequence of vertices displacements denoted as $\hat{D_i} = S2D(\hat{S_i})= \left(\hat{d_i}^0, \dots, \hat{d_i}^{K_i}\right) \in \mathbb{R}^{K_i \times M \times 3}$.
  \item The neutral 3D face $M_i^n$ is summed with each vertex displacement $\hat{D_i}$ to generate the 3D talking heads sequence denoted as $\hat{M_i} = \hat{D_i} + M_i^n$.
\end{enumerate}

\section{Experiments}
In this section, we present the experiments conducted to assess the efficacy of our proposed approach. Specifically, we conducted a comparative study with respect to two existing methods, namely, Faceformer~\cite{DBLP:journals/corr/abs-2112-05329} and VOCA~\cite{Cudeiro_2019_CVPR}. Our objective is to evaluate the performance of our approach relative to the state-of-the-art.

\subsubsection{VOCAset:}
Our experimental setup utilized the VOCAset, comprising of 12 actors, with an equal gender split of 6 males and 6 females. Each actor delivered 40 distinct sentences, with durations ranging from 3 to 5 seconds. The dataset includes high-fidelity audio recordings and 3D facial reconstructions per frame, captured at a frame rate of 60 fps.
The dataset was partitioned into three distinct subsets for the purposes of training, validation, and testing. The training subset consists of 8 actors, while the validation and test subsets include 2 actors each.

\subsection{Results}
To evaluate the efficacy of Faceformer and VOCA, we employed the pre-trained models made available by their respective authors. As Faceformer operates at a frame rate of 30 fps, we compared its output against the ground truth at this rate. Conversely, our approach and VOCA operate at a frame rate of 60 fps, and thus we compared their outputs against the original ground truth. Landmarks play a crucial role in evaluating the animation process, and their quality directly impacts the naturalness and realism of the resulting speech. To assess the effectiveness of landmark generation, we also evaluate the landmarks obtained from meshes generated by both Faceformer and VOCA.
All experiments were conducted exclusively on the test subset of the VOCAset dataset, and the presented results in Table~\ref{tab1} and Table~\ref{tab2} represent an average of all results.

\subsubsection{Lips Error:}
As suggested in~\cite{DBLP:journals/corr/abs-2112-05329}, we computed the Lips Error (LE) to assess the quality of the generated lip motion sequences. This metric is defined in~\cite{DBLP:journals/corr/abs-2112-05329} as the maximum of the $L_2$ distance between each lip vertex (or landmarks) of the generated sequence and those of the corresponding ground truth.

\subsubsection{Displacements Errors:}
We compared the displacement outputs generated by Faceformer and VOCA to those produced by our approach using ground truth data, as our model operates on the displacements. To evaluate the quality of the generated results, we utilized both cosine distance and $L_2$ distance metrics, specifically focusing on the average $L_2$ distance between all displacements (DE) and the maximum angle between all displacements (DAE). We calculated the error on both the landmark displacement outputs generated by S2L and those on vertices generated by concatenating S2L and S2D.
\begin{table}[h]
\centering
\caption{Comparison with the state-of-the-art. Errors are reported for Lips Error (LE) and Dense Error (DE) and displacement angle discrepancy (DAE) error.}\label{tab1}
\begin{adjustbox}{width=0.95\textwidth}
\begin{tabular}{c||c|c|c||c|c|c}
\hline
& \multicolumn{3}{c||}{\textbf{Landmarks}} & \multicolumn{3}{c}{\textbf{Dense}} \\
\hline
\hline
\textbf{Methods} &  \textbf{LE} (mm) & \textbf{DE} (mm) & \textbf{DAE} (Rad) &  \textbf{LE} (mm) & \textbf{DE} (mm) & \textbf{DAE} (Rad)\\
\hline
VOCA~\cite{Cudeiro_2019_CVPR} & 8.72 & 7.71 & 0.29 & 7.24 & 6.29 & 0.23\\
Faceformer~\cite{DBLP:journals/corr/abs-2112-05329} & 6.1 & 5.6 & 0.20 & 5.12 & 4.24 & 0.17\\
Ours & \textbf{5.01} & \textbf{4.42} & \textbf{0.13} & \textbf{4.31} & \textbf{3.42} & \textbf{0.12}\\
\hline
\end{tabular}
\end{adjustbox}
\end{table}
%\begin{table}[h]
%\centering
%\caption{Vertex Quantitative Results.}\label{tab1}
%\begin{tabular}{c|c|c|c}
%\hline
%Methods &  V-LE (mm) & V-DE (mm) & V-DAE (Rad)\\
%\hline
%VOCA & 0.7233 & 0.6247 & 0.234 \\
%Faceformer & 0.5128 & 0.4223 & 0.173 \\
%Ours & \textbf{0.4274} & \textbf{0.3452} & \textbf{0.109} \\
%\hline
%\end{tabular}
%\end{table}
Our proposed approach surpasses both Faceformer and VOCA in generating landmarks and vertices, as demonstrated in Table~\ref{tab1}. While the differences among models in terms of LE are insignificant, the gaps in DAE are more pronounced. This suggests that our approach produces more realistic and accurate landmark and vertex displacements, closely resembling ground truth data. This is unsurprising since our models were trained to minimize the angle between generated and ground truth displacements. Notably, the quantitative performance of landmark-based outcomes is inferior to that of vertex-based outcomes, which can be attributed to the relatively greater proportion of salient landmarks in speech compared to vertices in a mesh.
\begin{figure}[h!]
\centering
\includegraphics[width=0.9\textwidth]{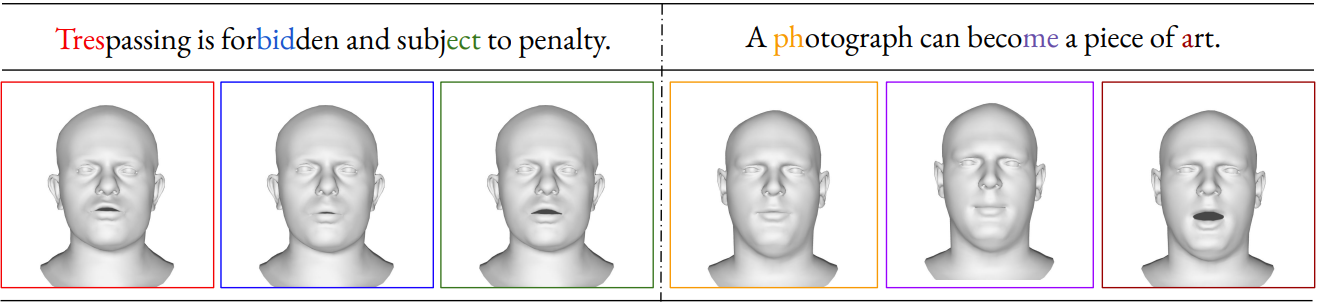}
\caption{Qualitative examples of the proposed framework.} \label{fig:qualitative}
\end{figure}
\begin{figure}[h!]
\centering
\includegraphics[width=0.9\textwidth]{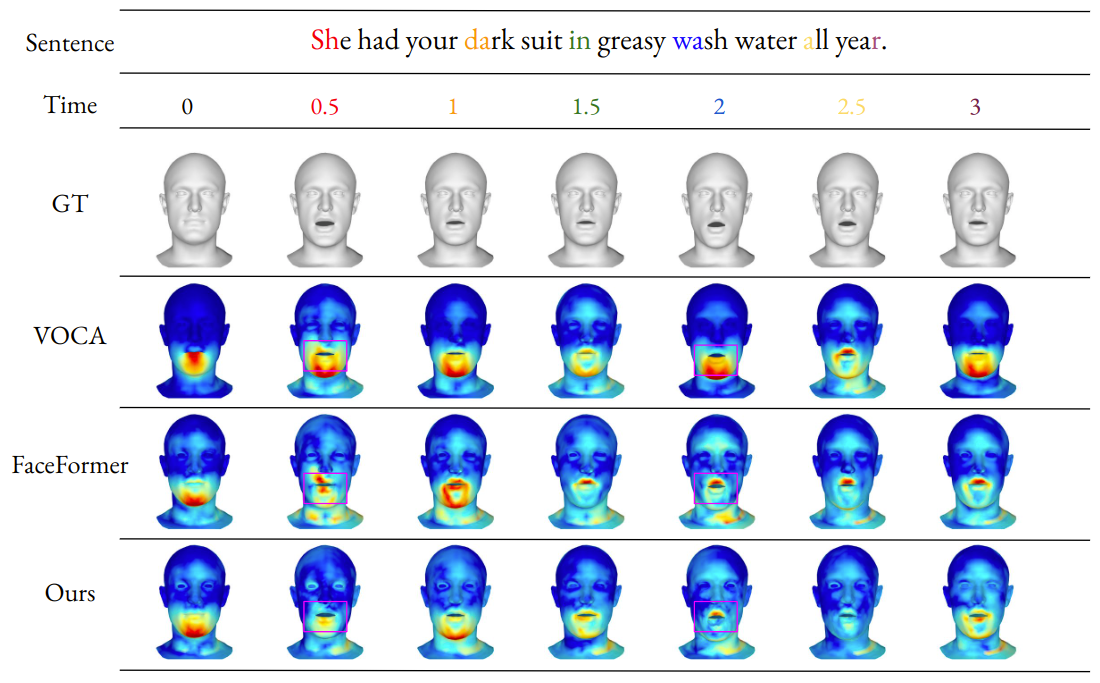}
\caption{Qualitative evaluation of our proposed framework in comparison to Faceformer and VOCA. The assessment is performed on generated meshes, where color gradation indicates the magnitude of the deviation from the groundtruth. Specifically, the color blue denotes a lower level of error, while the color red indicates a higher level of error.} \label{fig:heatmaps}
\end{figure}
Figure~\ref{fig:qualitative} illustrates two qualitative examples of our proposed framework. The effectiveness of our framework is evident in accurately capturing the lip closure during the pronunciation of consonants such as \textit{``b'' }, \textit{``p''}, and \textit{``m''}.
Figure~\ref{fig:heatmaps} shows a comparison of the meshes generated by VOCA, Faceformer, and our model, which outperforms other methods mostly when the mouth takes certain positions during speech, such as when generating displacements for phonemes like \textit{``Sh''}, \textit{``Wa''} or \textit{``Gl''}. 
Additional quantitative results and comparisons can be found in the supplementary video.

\subsection{Ablation study}
%In the outset of our study, we attempted to employ a non-pretrained model for the purpose of extracting information from audio spectrograms. However, that model was not able to  generalize in a proper way, leading to suboptimal outcomes. To address this limitation, we advocate the adoption of a pre-trained audio model, which can endow our methodology with superior generalization capabilities. In particular, the use of a pre-trained audio model empowers the production of language-agnostic audio-driven animations.

In order to evaluate the efficacy of our proposed approach, we performed an ablation study on the selection of loss functions employed during model training. Our utilized loss functions are widely accepted in the field and have been employed in previous works, in contrast to our novel introduction of the cosine loss. Thus, to gauge the enhancement provided by the latter, we compared the errors obtained from models trained both with and without the cosine loss, utilizing the previously defined metrics for landmarks and vertices displacements.

%\begin{table}[h]
%\centering
%\caption{Landmarks ablation study.}\label{tab3}
%\begin{tabular}{c|c|c|c}
%\hline
%Loss &  L-LE (mm) & L-DE (mm) & L-DAE (Rad)\\
%\hline
%$L_{S2L}$ w/o $L_{cos}$ & 0.5382 & 0.4662 & 0.187 \\
%$L_{S2L}$  & \textbf{0.5035} & \textbf{0.4384} & \textbf{0.132} \\
%\hline
%\end{tabular}
%\end{table}

%\begin{table}[h]
%\centering
%\caption{Vertex ablation study.}\label{tab4}
%\begin{tabular}{c|c|c|c}
%\hline
%%Loss &  V-LE (mm) & V-DE (mm) & V-DAE (Rad)\\
%\hline
%$L_{S2L}$ and $L_{S2D}$ w/o $L_{cos}$ & 0.4441 & 0.3667 & 0.169 \\
%$L_{S2L}$ and $L_{S2D}$ & \textbf{0.4274} & \textbf{0.3452} & %\textbf{0.109} \\
%\hline
%\end{tabular}
%\end{table}

\begin{table}[h]
\centering
\caption{Ablation study of the Cosine Loss. Errors are reported for Lips Error (LE), Dense Error (DE), and displacement angle discrepancy (DAE) error.}\label{tab2}
\begin{adjustbox}{width=0.95\textwidth}
\begin{tabular}{c||c|c|c||c|c|c}
\hline
& \multicolumn{3}{c||}{\textbf{Landmarks}} & \multicolumn{3}{c}{\textbf{Dense}} \\
\hline
\hline
\textbf{Loss} &  \textbf{LE} (mm) & \textbf{DE} (mm) & \textbf{DAE} (Rad) &  \textbf{LE} (mm) & \textbf{DE} (mm) & \textbf{DAE} (Rad)\\
\hline
w/o $L_{cos}$ & 5.35 & 4.67 & 0.19 & 4.48 & 3.62 & 0.17\\
w/ $L_{cos}$ & \textbf{5.01} & \textbf{4.42} & \textbf{0.13} & \textbf{4.32} & \textbf{3.42} & \textbf{0.12}\\
\hline
\end{tabular}
\end{adjustbox}
\end{table}

\noindent
According to Table~\ref{tab2}, incorporating the cosine loss during training of the two models enhances the fidelity of the generated displacements for both landmarks and vertices. As a result, the utilization of the cosine loss augments the potential of our framework to produce convincing talking heads. 
Additional quantitative results about the advantage of cosine loss usage can be found in the supplementary video.

\subsection{Limitations}
While this framework yields reasonably accurate animations, it is not without limitations. The primary challenge is the deficiency in the expressive capacity of the generated meshes, which lack emotional nuances due to the training data inexpressive nature. A possible step forward is to enhance the realism of the animation by modeling both expressions of emotion and deformations of the upper part of the face.
Furthermore, our model's generation times, though lower than those of other techniques like VOCA or Faceformer, remain inadequate for real-time applications. %Furthermore, the mesh topology used to train the second model is fixed and does not allow for mesh densification in different topologies. To overcome this limitation, the use of landmarks can prove useful. In our future work, we will explore this avenue.

\section{Conclusions}
In this paper, we have introduced a new approach for generating 3D talking heads based on raw audio inputs. Our experimental results indicate that capturing the motion of facial landmarks is sufficient to effectively represent speech movements. Additionally, training two separate models to separate this motion from the movement of mesh vertices leads to improved realism and accuracy in lip movements.
However, the generation of 3D facial animations raises ethical concerns. Creating fabricated narratives using generated 3D faces can be risky and have both intentional and unintentional consequences for individuals and society as a whole. It is important to emphasize that technology should always prioritize human-centered considerations. Therefore, it is crucial to carefully consider the social and psychological impacts of such technology.
%Ignoring the ethical implications of creating technology is hazardous.

\bibliographystyle{splncs04}
\bibliography{samplepaper}

\begin{thebibliography}{10}
\providecommand{\url}[1]{\texttt{#1}}
\providecommand{\urlprefix}{URL }
\providecommand{\doi}[1]{https://doi.org/#1}

\bibitem{DBLP:journals/corr/abs-2006-11477}
Baevski, A., Zhou, H., Mohamed, A., Auli, M.: wav2vec 2.0: {A} framework for
  self-supervised learning of speech representations. CoRR
  \textbf{abs/2006.11477} (2020)

\bibitem{DBLP:journals/corr/abs-1905-02876}
Bouritsas, G., Bokhnyak, S., Ploumpis, S., Bronstein, M.M., Zafeiriou, S.:
  Neural 3d morphable models: Spiral convolutional networks for 3d shape
  representation learning and generation. CoRR  \textbf{abs/1905.02876} (2019)

\bibitem{DBLP:journals/corr/ChungJZ17}
Chung, J.S., Jamaludin, A., Zisserman, A.: You said that? CoRR
  \textbf{abs/1705.02966} (2017)

\bibitem{Cudeiro_2019_CVPR}
Cudeiro, D., Bolkart, T., Laidlaw, C., Ranjan, A., Black, M.J.: Capture,
  learning, and synthesis of 3d speaking styles. In: Proceedings of the
  IEEE/CVF Conference on Computer Vision and Pattern Recognition (CVPR) (June
  2019)

\bibitem{DEMARTINO2006971}
{De Martino}, J.M., {Pini Magalhães}, L., Violaro, F.: Facial animation based
  on context-dependent visemes. Computers \& Graphics  \textbf{30}(6),
  971--980 (2006). \doi{https://doi.org/10.1016/j.cag.2006.08.017},
  \url{https://www.sciencedirect.com/science/article/pii/S0097849306001518}

\bibitem{10.1145/2897824.2925984}
Edwards, P., Landreth, C., Fiume, E., Singh, K.: Jali: An animator-centric
  viseme model for expressive lip synchronization. ACM Trans. Graph.
  \textbf{35}(4) (jul 2016). \doi{10.1145/2897824.2925984},
  \url{https://doi.org/10.1145/2897824.2925984}

\bibitem{DBLP:journals/corr/abs-2112-05329}
Fan, Y., Lin, Z., Saito, J., Wang, W., Komura, T.: Faceformer: Speech-driven 3d
  facial animation with transformers. CoRR  \textbf{abs/2112.05329} (2021)

\bibitem{982373}
Kalberer, G., Van~Gool, L.: Face animation based on observed 3d speech
  dynamics. In: Proceedings Computer Animation 2001. Fourteenth Conference on
  Computer Animation (Cat. No.01TH8596). pp. 20--251 (2001).
  \doi{10.1109/CA.2001.982373}

\bibitem{10.1145/3072959.3073658}
Karras, T., Aila, T., Laine, S., Herva, A., Lehtinen, J.: Audio-driven facial
  animation by joint end-to-end learning of pose and emotion. ACM Trans. Graph.
   \textbf{36}(4) (jul 2017). \doi{10.1145/3072959.3073658},
  \url{https://doi.org/10.1145/3072959.3073658}

\bibitem{otberdout2022sparse}
Otberdout, N., Ferrari, C., Daoudi, M., Berretti, S., Del~Bimbo, A.: Sparse to
  dense dynamic 3d facial expression generation. In: Proceedings of the
  IEEE/CVF Conference on Computer Vision and Pattern Recognition. pp.
  20385--20394 (2022)

\bibitem{otberdout2023generating}
Otberdout, N., Ferrari, C., Daoudi, M., Berretti, S., Del~Bimbo, A.: Generating
  multiple 4d expression transitions by learning face landmark trajectories.
  IEEE Transactions on Affective Computing  (2023)

\bibitem{DBLP:journals/corr/abs-2104-08223}
Richard, A., Zollh{\"{o}}fer, M., Wen, Y., la~Torre, F.D., Sheikh, Y.:
  Meshtalk: 3d face animation from speech using cross-modality disentanglement.
  CoRR  \textbf{abs/2104.08223} (2021)

\bibitem{10.1145/3072959.3073640}
Suwajanakorn, S., Seitz, S.M., Kemelmacher-Shlizerman, I.: Synthesizing obama:
  Learning lip sync from audio. ACM Trans. Graph.  \textbf{36}(4) (jul 2017).
  \doi{10.1145/3072959.3073640}, \url{https://doi.org/10.1145/3072959.3073640}

\bibitem{NIPS2017_3f5ee243}
Vaswani, A., Shazeer, N., Parmar, N., Uszkoreit, J., Jones, L., Gomez, A.N.,
  Kaiser, L.u., Polosukhin, I.: Attention is all you need. In: Guyon, I.,
  Luxburg, U.V., Bengio, S., Wallach, H., Fergus, R., Vishwanathan, S.,
  Garnett, R. (eds.) Advances in Neural Information Processing Systems.
  vol.~30. Curran Associates, Inc. (2017),
  \url{https://proceedings.neurips.cc/paper1\_files/paper/2017/file/3f5ee243547dee91fbd
  053c1c4a845aa\-Paper.pdf}

\bibitem{DBLP:journals/corr/abs-1906-06337}
Vougioukas, K., Petridis, S., Pantic, M.: Realistic speech-driven facial
  animation with gans. CoRR  \textbf{abs/1906.06337} (2019)

\bibitem{DBLP:journals/corr/abs-2004-12992}
Zhou, Y., Li, D., Han, X., Kalogerakis, E., Shechtman, E., Echevarria, J.:
  Makeittalk: Speaker-aware talking head animation. CoRR
  \textbf{abs/2004.12992} (2020)

\end{thebibliography}

\end{document}